\documentclass[letterpaper, 10 pt, conference]{ieeeconf}  

\IEEEoverridecommandlockouts                              

\usepackage{amsmath} 
\usepackage{amssymb}  
\usepackage{graphicx}
\usepackage{siunitx}

\title{\LARGE \bf
Measurement and Potential Field-Based Patient Modelling for Model-Mediated Tele-ultrasound
}

\author{Ryan~S.~Yeung$^{1}$, David~G.~Black$^{2}$, and~Septimiu~E.~Salcudean$^{2}$,~\IEEEmembership{Life~Fellow,~IEEE}%
\thanks{$^{1}$School of Biomedical Engineering, University of British Columbia, Vancouver, Canada
        {\tt\small ryeung1@student.ubc.ca}}%
\thanks{$^{2}$Department of Electrical and Computer Engineering, University of British Columbia, Vancouver, Canada
        {\tt\small \{dgblack,tims\}@ece.ubc.ca}}%
}

\begin{document}

\maketitle
\thispagestyle{empty}
\pagestyle{empty}

\begin{abstract}
Teleoperated ultrasound can improve diagnostic medical imaging access for remote communities. Having accurate force feedback is important for enabling sonographers to apply the appropriate probe contact force to optimize ultrasound image quality. However, large time delays in communication make direct force feedback impractical. Prior work investigated using point cloud-based model-mediated teleoperation and internal potential field models to estimate contact forces and torques. We expand on this by introducing a method to update the internal potential field model of the patient with measured positions, forces and torques for more transparent model-mediated tele-ultrasound. We first generate a point cloud model of the patient's surface and transmit this to the sonographer in a compact data structure. This is converted to a static voxelized volume where each voxel contains a potential field value. These values determine the forces and torques, which are rendered based on overlap between the voxelized volume and a point shell model of the ultrasound transducer. We solve for the potential field using a convex quadratic that combines the spatial Laplace operator with measured forces and torques. This was evaluated on volunteers ($n=4$) by assessing the accuracy of rendered forces and torques. Results showed the addition of measurements to the model reduced the force magnitude RMSE by an average of 7.42~N, the force vector angle error by an average of 3.71$^{\circ}$, and the torque vector angle error by an average of 64.0$^{\circ}$ compared to using only Laplace's equation.
\end{abstract}

\section{INTRODUCTION}
Despite the importance of medical ultrasound (US) for patient care, many rural and remote communities still lack local access, forcing them to endure lengthy and costly travel for these services \cite{adams_access_2021}. The emergence of portable US presents a potential solution for this problem. However, the lack of experienced US users in these low-resource environments is still a major obstacle \cite{tang_portable_2022}.

To overcome this obstacle, advances have been made in robotic US technology to enable either autonomous US imaging or teleoperated US \cite{jiang_robotic_2023}. The former has recently garnered increased interest due to advances in machine learning; however, it still faces many challenges that limit its translational feasibility. As such, teleoperated US remains an important area of study. Since early implementations of teleoperated US \cite{salcudean_robot_2000}, many advances have been made, including the development of human teleoperation, a low-cost mixed-reality system in which a novice person is controlled as a cognitive robot \cite{black_human_2024} by the sonographer. In teleoperated US, the sonographer typically manipulates a haptic device to input their desired trajectory and receive force feedback. This bilateral teleoperation approach relies on two-way communication, so network delays caused by large distances separating the sonographer and follower robot or novice person can degrade the system stability, transparency and overall performance \cite{franken_delay_2011}. 

An alternative approach is model-mediated teleoperation (MMT) in which the remote environment is modeled on the local side, and force feedback is generated through interactions with the model \cite{xu_mmt_2016}. The use of MMT for teleoperated abdominal US was demonstrated in \cite{yeung_mixed_2025}, in which an ellipsoid was used to model the patient's torso based on measured parameters. Another method of modelling the remote environment is to generate a point cloud using a time-of-flight (TOF) depth camera \cite{xu_point_2014}. TOF depth cameras are widely used in robotic teleoperation and can also be found in mixed-reality headsets such as the Magic Leap 2 (Magic Leap Inc., Florida, USA). 

Using point cloud-based MMT is beneficial because it can render complex object geometry and reduce computational complexity by omitting the conversion into a 3D mesh. Rendering haptics directly from a point cloud was initially demonstrated in three degrees-of-freedom (3-DOF) \cite{ryden_proxy_2013}. In this method, a proxy is tied to the haptic interface point, (the position of the end effector in virtual space), by a virtual spring-damper, and the proxy is constrained above a surface estimated from a set of points in the point cloud. This was later extended to six degrees-of-freedom (6-DOF), where a generalized constrained acceleration for the virtual tool was determined based on the point cloud, and the force and torque were computed based on the difference between the virtual tool and haptic device configurations \cite{ryden_contraint_2013}.

\begin{figure*}[t]
    \centering
    \includegraphics[width=0.8\textwidth]{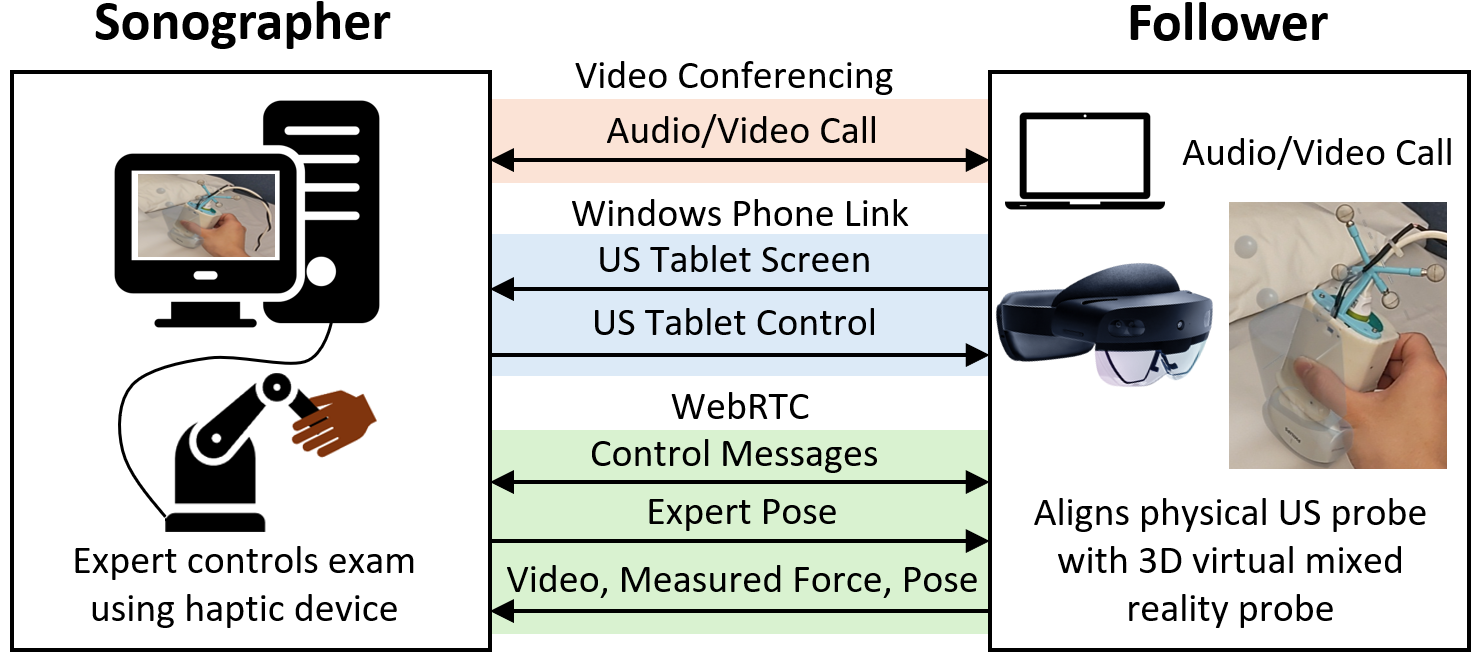}
    \caption{Overview of the human teleoperation system.}
    \label{fig:humanteleop}
\end{figure*}

Voxmap Pointshell (VPS) is another model-based haptic rendering method, in which manipulated objects are represented as point shells (points with unit normal vectors), and static rigid objects are represented as a voxel map. Each voxel contains a 2-bit value indicating whether it is free space, interior, surface or proximity. Force and torque are rendered for each manipulated object at each haptic frame by the interpenetration of the points with the voxels \cite{mcneely_voxel_1999}. Iterations of this method have since been explored, such as using signed distance fields to represent the static objects \cite{xu_signed_2017}. These methods are fast and robust for preventing interpenetration of geometrically complex rigid objects.

Another fast approach to simulating contact forces is through Pressure Field Contact \cite{elandt_pressure_2019}. This method predicts the contact surface and net contact wrench between two objects. A pressure field is precomputed for the interior of each object by solving Laplace's equation. When the objects overlap, a contact surface is identified where the pressure fields are equal. Tractions on the contact surface are integrated to determine the contact wrench between the objects. This fast and continuous method of computing net contact wrench is particularly useful in simulating robot manipulation for learning.

While these methods are effective and fast at rendering force feedback, they do not address the challenge of modelling the physical properties of the patient. Appropriate probe contact force is critical for US image quality, and sonographers typically regulate this using force cues \cite{suchon_presssure_2025}. In prior work, recursive least squares were used to estimate parameters of the Kelvin-Voigt or Hunt-Crossley force models for impedance estimation, but practical applications of this are mostly limited to 1-DOF \cite{diolaiti_impedance_2005,fu_predictive_2023,batty_transparent_2022}. In \cite{xu_point_2014}, Xu et al. described a point cloud-based MMT in which the friction coefficient and stiffness were estimated by measuring the forces and follower end-effector positions. However, this approach still required continuously sending the parameters to the expert operator to update the model and did not capture spatial variations of the physical properties.

In this paper, we propose a model-mediated teleoperation method for teleoperated abdominal US that incorporates measured position, forces, and torques to estimate stiffness. Inspired by \cite{mcneely_voxel_1999} and \cite{elandt_pressure_2019}, we use a point cloud surface and internal patient model initialized by Laplace's equation. However, unlike in prior approaches, we update the volumetric Laplace's equation with measurements of positions, forces, and torques, adjusting the internal potential field model accordingly. This enables more accurate rendering of the stiffness map, incorporating local variations. In this manuscript, we first describe how the patient model is initialized and the method for rendering forces and torques from the model. We then describe how the stiffness is estimated and mapped to the patient model. Finally, we describe the experiments conducted to quantify the force and torque accuracy to assess the transparency.

\section{METHODOLOGY}
\subsection{Human Teleoperation}
The human teleoperation system described in references \cite{black_human_2024,black_mixed_2024,black_robot_2023} is the subject of current research. In order to better place our work in context, we briefly outline the main concepts and instrumentation used. With reference to Figure~\ref{fig:humanteleop}, human teleoperation enables tightly-coupled remote guidance by a sonographer of a human follower, who acts as a flexible, cognitive robot. The follower wears a mixed-reality headset such as the Microsoft HoloLens~2 (Microsoft, Redmond, WA) or the Magic Leap~2, which projects a 3D virtual guiding US transducer in their environment. The follower aligns the real US transducer with the virtual transducer controlled remotely by the sonographer as if it is the end-effector of a robotic arm. The sonographer uses a haptic device, such as the Touch X (3D Systems, Rock Hill, SC) to input the desired pose and motion of the virtual transducer.

All communication occurs over the Internet using Web Real Time Communication (WebRTC) as described in \cite{black_communication_2024}. The real-time feeds of the US image and point-of-view video from the follower's headset are sent to the sonographer. The US transducer is additionally instrumented with a force sensing shell \cite{black_force_2024} and infrared markers to enable measurement of force and pose respectively \cite{black_pose_2024}.

\subsection{Patient Surface Extraction}\label{sec:surf_extract}
\begin{figure}[t]
    \centering
    \includegraphics[width=\linewidth]{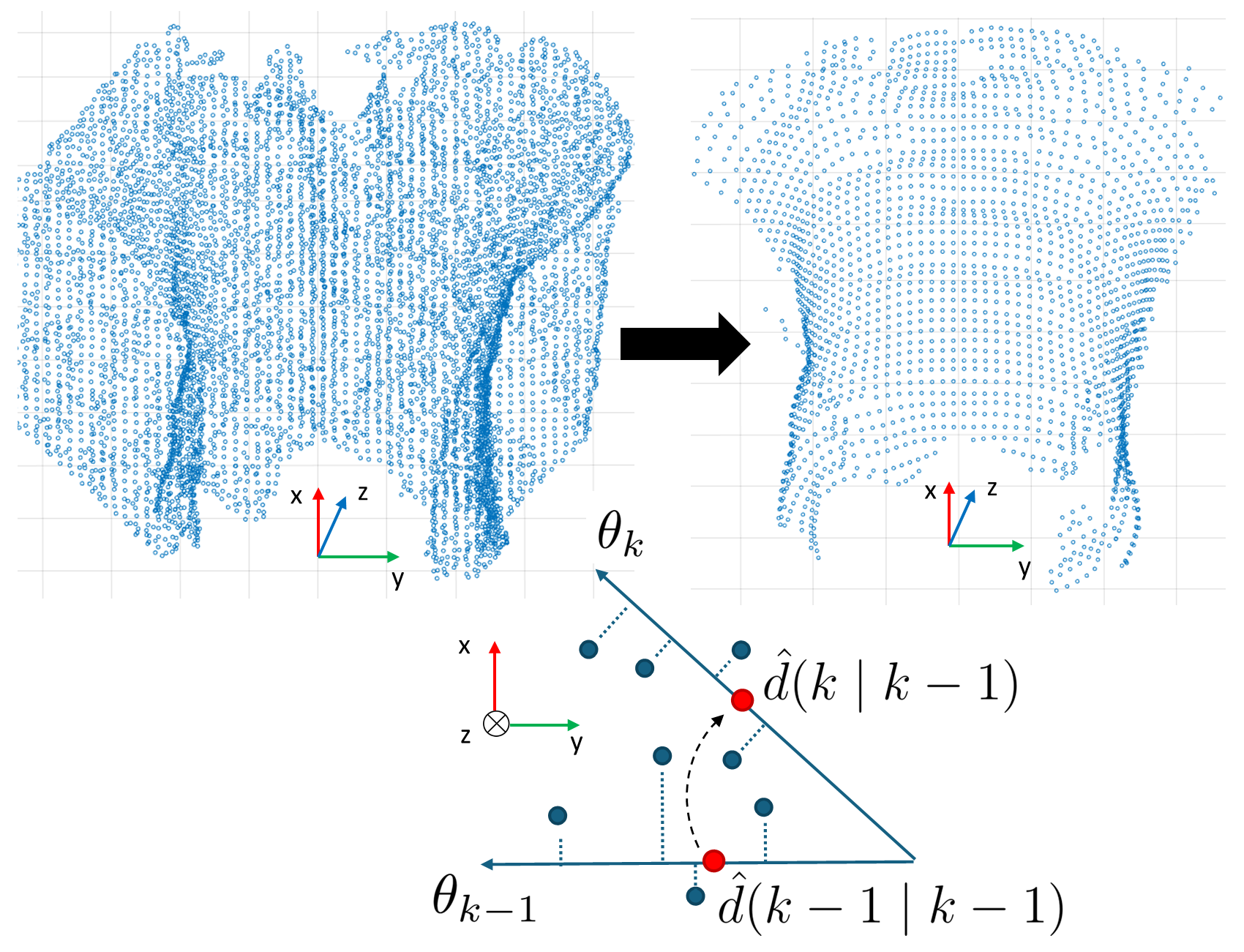}
    \caption{Overview of patient surface extraction. The raw point cloud (left) is converted to a structured point cloud representation of the patient surface in cylindrical coordinates (right). The bottom middle image shows an example of the contour extraction method where the blue circles are the candidate surface points and the red circles are the estimated surface points.}
    \label{fig:pointcloud}
\end{figure}
A TOF depth camera is first used to acquire a point cloud measurement of the patient's torso. Multiple scans from different angles are captured and stitched together by measuring the relative pose of the camera for each scan. We use a contour extraction algorithm adapted from \cite{Abolmaesumi_Sirouspour_Salcudean_2002} to convert the raw point cloud measurement into a structured point cloud representation of the patient surface in cylindrical coordinates, as illustrated in Figure~\ref{fig:pointcloud}. We summarize this method below.

First, an interior central z-axis of the patient is estimated as the longitudinal axis of the US probe offset downwards from its tip as it is placed longitudinally above the patient's sternum. The dimension along the z-axis is discretized into equally spaced regions, which we will refer to as slices. Within each slice, $R$ angularly equispaced vectors are projected outward from the axis in the radial direction. All points in the slice and within a threshold distance from the projected vector are taken as candidate surface points. Let $r_i(k)$ be the radius of the $i^{\text{th}}$ candidate surface point projected onto vector $k$. A Kalman filter is used to update the estimated radius based on the radius estimate at the previous angle and the current candidate surface points. Let the radius of the surface point along vector $k$ be $\hat{d}(k|k)$, and $\hat{d}(k|k-1)$ be the predicted state at iteration $k$. Then,
\begin{equation}
    \hat{d}(k|k) = \hat{d}(k|k-1) + W(k)(z(k) - \hat{d}(k|k-1)),
\end{equation}
where $W(k)$ is the Kalman filter gain and $z(k)$ is an unknown noisy version of $d(k)$. As measurement into this Kalman filter, we use
\begin{equation}
    y(k) = \sum^M_{i=1}r_i(k)\beta_i(k),
\end{equation}
where
\begin{equation}
    \beta_i(k) = \frac{p_i(k)}{\sum_ip_i(k)},
\end{equation}
and
\begin{equation}
    p_i(k) = \frac{1}{\sqrt{2\pi S(k)}}e^{-\frac{(y(k)-\hat{d}(k|k-1))^2}{2S(k)}}D(r_i(k)),
\end{equation}
where
\begin{equation}
    D(r_i(k)) = \left ( 1 - \frac{r_i(k) - r_{min}(k)}{r_{max}(k) - r_{min}(k)} \right )^2
\end{equation}
$S(k)$ is the variance of the candidate surface points. The function $D(r_i(k))$ produces an exponentially decreasing inverse rank weight so that candidate surface points with a larger radius have a smaller probability of being the correct measurement. This is to reduce the effect of unwanted points captured of the bed surface when scanning the patient. This is repeated for all slices along the z-axis. An additional benefit of converting the raw point cloud into a structured point cloud is that it becomes more efficient to send. If the structured point cloud has $M_z$ slices, then the entire point cloud can be sent as a $M_\theta \times M_z$ matrix containing the estimated radius for each corresponding angle and slice pair. The only additional information required are the positions of the first and last slices and angles, and the constant step size between each slice and angle.

\subsection{Force Rendering}\label{sec:frce_rend}
\begin{figure}[t]
    \centering
    \includegraphics[width=0.5\linewidth]{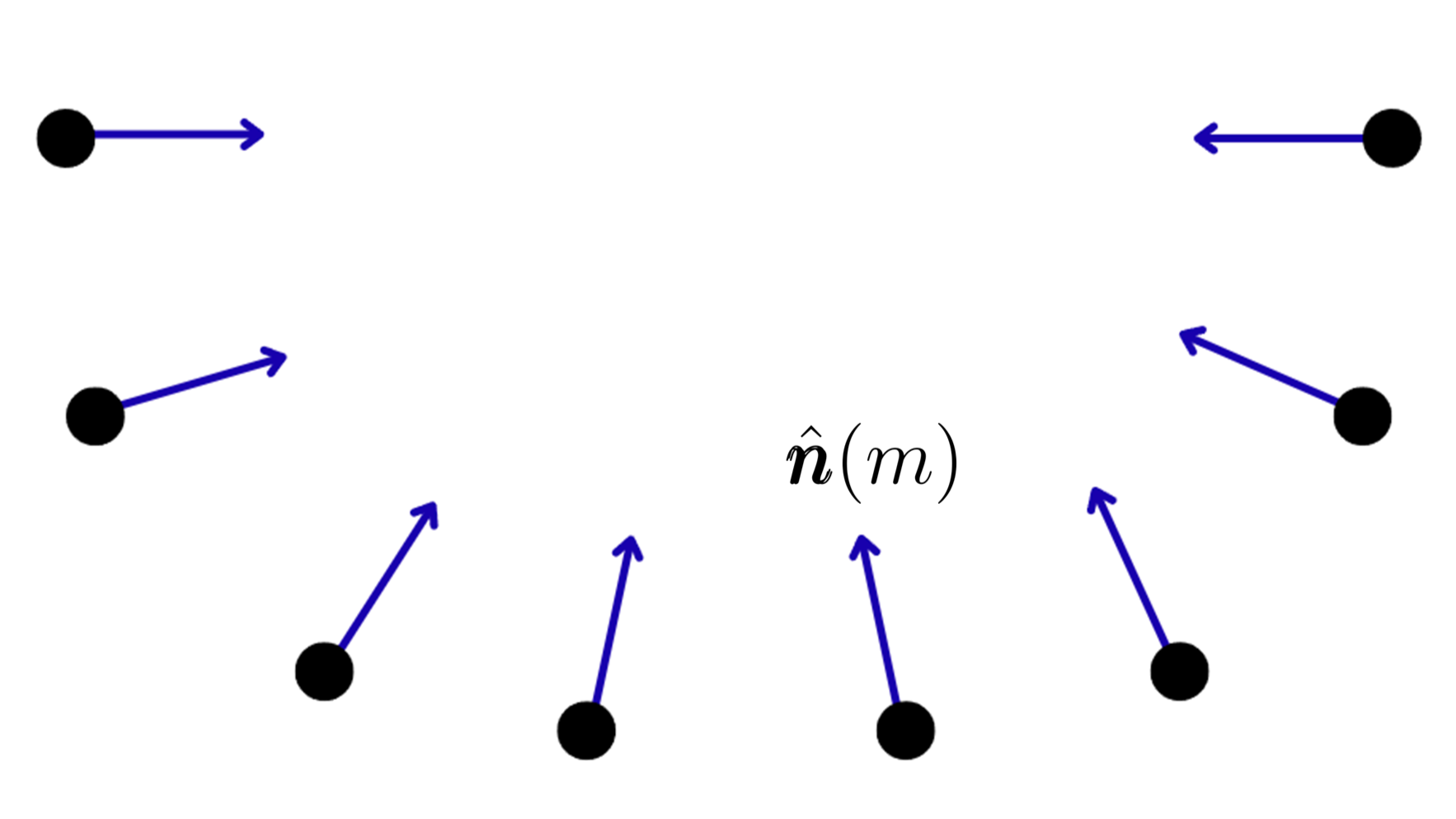}
    \caption{An example point shell in 2D with $M_m = 8$.}
    \label{fig:pointshell}
\end{figure}
\begin{figure}[t]
    \centering
    \includegraphics[width=0.9\linewidth]{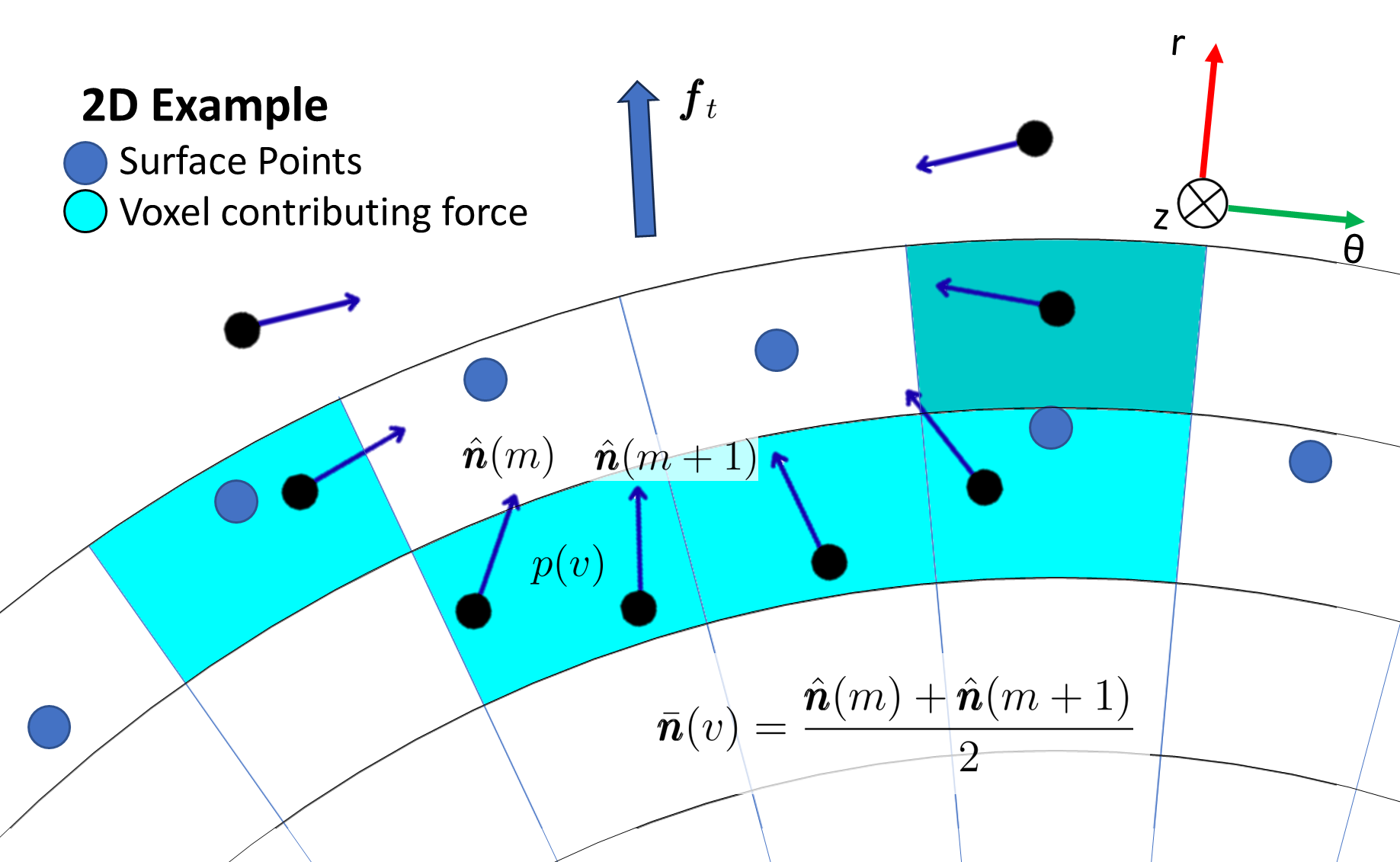}
    \caption{Transverse slice showing the force rendering method in 2D. The blue dots represent the surface points and the black dots represent the point shell. Voxels shaded light blue contribute to the force rendering and the darker shaded voxel has an overlapping point, but does not contribute to the force rendering because the point is outside of the surface.}
    \label{fig:voxelgrid}
\end{figure}
We use a force rendering model similar to the Voxmap Pointshell haptic rendering method \cite{mcneely_voxel_1999}. We model the patient as a static voxmap, generated by discretizing the volume contained by the patient surface points into $M_v$ voxels in cylindrical coordinates with constant step size in $r$, $z$, and $\theta$, so $M_v = M_r \times M_z \times M_\theta$. Each voxel has a unique index $v \in \{1,2,...,M_v\}$. We define the potential field as a vector $\pmb{p}\in \mathbb{R}^{M_v \times 1}$, so each voxel has a corresponding potential $p(v)$. The US probe is represented as a dynamic point cloud with $M_m$ points, each with a corresponding inward pointing surface unit vector normal $\hat{\pmb{n}}(m)$, $m \in \{1,2,...,M_m\}$. We refer to the combined point cloud and unit vector normals as a point shell (Figure~\ref{fig:pointshell}). 

The pose of the point shell is coupled to the sonographer's commanded pose, i.e., that of the virtual transducer. Let us consider a single timestep $t$. In this timestep, we query the voxel that each point $m$ lies in, denoted as $v^{(m)}$. The process has linear time complexity in $M_m$ since the voxel locations are known, so it involves a simple indexing operation for each point. Only points penetrating into the patient surface, i.e. radial position less than the patient surface radius at that angular location, are considered. Figure~\ref{fig:voxelgrid} shows a 2D example of a point shell penetrating into the patient surface (voxels shaded blue). The darker shaded voxel contains a point, but its radial position is further than the surface point at that angular location so it is ignored. After $v^{(m)}$ has been determined for all $M_m$ points, each voxel will have an associated set of points that lie inside it. For voxel $v$, we take the average of the normal vectors of these points, which we denote $\bar{\pmb{n}}(v)$. The force contribution of $v$ is then computed as
\begin{equation}\label{eq:fnp}
    \pmb{f}(v) = \bar{\pmb{n}}(v)p(v)
\end{equation}
The total force on the point shell probe at this timestep $t$ is the sum of these force contributions across all voxels
\begin{equation}\label{eq:fNP}
    \pmb{f}_t = \sum^{M_v}_{v=1} \bar{\pmb{n}}(v)p(v) = N_t\pmb{p}
\end{equation}
The matrix $N_t \in \mathbb{R}^{3 \times M_v}$ is sparse since most voxels will not have probe points inside it, making $\bar{\pmb{n}}(v) = \pmb{0}_3$. A complete set of measurements is then defined by stacking $N_t$ and $\pmb{f}_t$ at multiple times row-wise, leading to $\pmb{f} = N\pmb{p}$, where $\pmb{f} \in \mathbb{R}^{3M_t \times 1}$ and $N \in \mathbb{R}^{3M_t \times M_v}$ for $M_t$ total measurements.

\subsection{Torque Rendering}
To render torques, we first precompute the moment arm for every point in the probe point shell relative to the tip of the probe. We denote this as $\pmb{r}(m)$ for point $m$. We then form the vector $\pmb{w}(m) = \pmb{r}(m)\times\hat{\pmb{n}}(m)$ for each point. As described previously, at a single timestep $t$, we determine the set of points that lie inside each voxel. For voxel $v$, we take the average of $\pmb{w}(m)$ for all $m$ that lie inside the voxel and denote this $\bar{\pmb{w}}(v)$. The torque contribution of $v$ is then computed as
\begin{equation}
    \pmb{\tau}(v) = \bar{\pmb{w}}(v)p(v),
\end{equation}
and the total torque on the point shell probe at this timestep $t$ is the sum
\begin{equation}\label{eq:tWP}
    \pmb{\tau}_t = \sum^{M_v}_{v=1}\bar{\pmb{w}}(v)p(v) = W_t\pmb{p}
\end{equation}
Once again, $W_t \in \mathbb{R}^{3 \times M_v}$ is sparse since most voxels will not have probe points inside. A complete set of measurements is then defined by stacking $W_t$ and $\pmb{\tau}_t$ at multiple times row-wise, leading to $\pmb{\tau} = W\pmb{p}$, where $\pmb{\tau} \in \mathbb{R}^{3M_t \times 1}$ and $W \in \mathbb{R}^{3M_t \times M_v}$ for $M_t$ total measurements.

\subsection{Stiffness Estimation}\label{sec:estimation_algorithm}
\begin{figure}[t]
    \centering
    \includegraphics[width=0.8\linewidth]{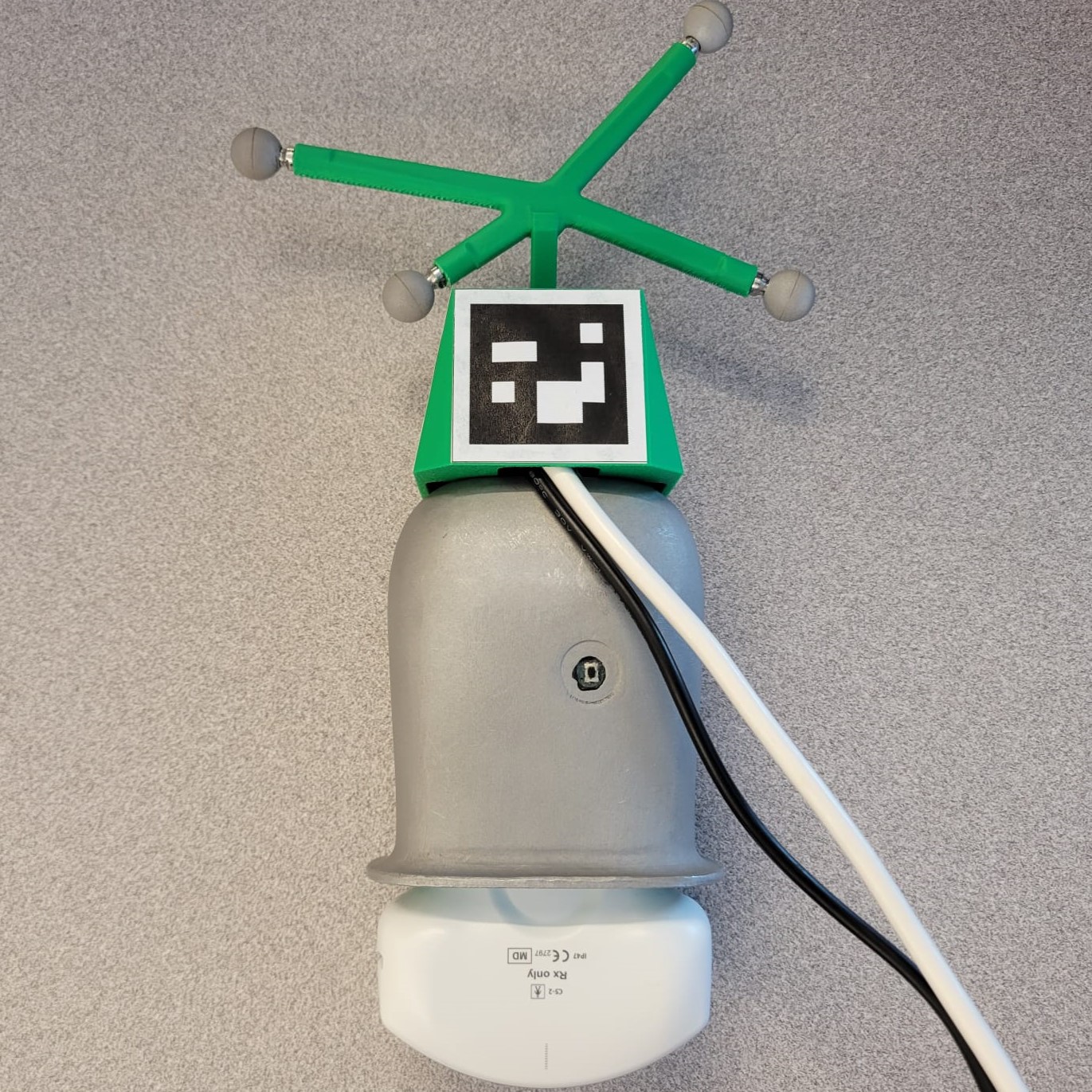}
    \caption{Philips Lumify US transducer instrumented with a force sensor shell, AruCo markers, and IR reflective markers.}
    \label{fig:probe}
\end{figure}
To determine the values $p(v)$, we treat the interior of the patient as a potential field, modeled using Laplace's equation, similar to \cite{elandt_pressure_2019} but in cylindrical coordinates:
\begin{equation}\label{eq:laplace}
    \nabla^2p = \frac{1}{r}\frac{\partial p}{\partial r} + \frac{\partial^2 p}{\partial r^2} + \frac{1}{r^2}\frac{\partial^2p}{\partial\theta^2} + \frac{\partial^2p}{\partial z^2} = 0
\end{equation}
We use Dirichlet boundary conditions and set the inner radial boundary to a positive value and the outer radial boundary to a negative value. At the radial positions of the patient surface points, we set the boundary condition to zero. At $z=z_{min}$ and $z=z_{max}$, we solve Laplace's equation in polar coordinates using the radial boundary conditions described above, and use the solution as boundary conditions in $z$ to then solve Laplace's equation in cylindrical coordinates.

We solve Laplace's equation numerically over the discrete voxmap by approximating each derivative using the central finite difference approximation. This allows us to form the matrix $\mathcal{L} \in \mathbb{R}^{M_v \times M_v}$ and solve $\mathcal{L}\pmb{p} = \pmb{b}$ where $\pmb{b}=\pmb{0}_{M_v}$ except for at boundary conditions. This results in a solution where $p(v)>0$ for voxels inside the patient and $p(v)<0$ for voxels outside the patient. This allows for fast collision detection by simply checking the sign of $p(v)$ and ignoring values less than zero.
While Laplace's equation can model the increasing stiffness of the patient for deeper penetration under the surface, it is not based on any real measurements and cannot capture variations across different parts of the torso. As such, we inform this model with real measurements of the forces and torques to improve its accuracy. To do this, we include Equations \ref{eq:fNP} and \ref{eq:tWP} as regularized terms in the objective:
\begin{equation}\label{eq:objective}
    \min_{\pmb{p}} ||\mathcal{L}\pmb{p}-\pmb{b}||^2 + \lambda||N\pmb{p}-\pmb{f}||^2 + \lambda||W\pmb{p}-\pmb{\tau}||^2
\end{equation}
With this approach, the smoothness of the potential field modeled by Laplace's equation is conserved while also incorporating known information from forces and torques measured during a scan. The regularization constant $\lambda$ was chosen through grid search on data collected from a phantom and a value of $\num{1e-4}$ was chosen. Minimizing this objective can be expressed as solving the following:
\begin{equation}\label{eq:Qpd}
    Q\pmb{p}=\pmb{d},
\end{equation}
where
\begin{align}
    Q &= L^TL + \lambda N^TN + \lambda W^TW \label{eq:Q} \\
    \pmb{d} &= L^T\pmb{b} + \lambda N^T\pmb{f} + \lambda W^T\pmb{\tau} \label{eq:d}
\end{align}
In general, $Q$ is symmetric and positive semidefinite. However, since we generate $L$ with Dirichlet boundary conditions, $L$ is full column rank. This makes $Q$ positive definite regardless of the properties of $N$ and $W$, guaranteeing a unique solution.

\subsection{Experiments}
We conducted experiments to compare the accuracy of rendered forces and torques between the model with force/torque information and without. For simplicity, we will refer to the model with force/torque information as the force-informed model throughout the remainder of the manuscript. The patient surface extraction method described in Section \ref{sec:surf_extract} was implemented on the Magic Leap 2, leveraging the built-in TOF depth camera and head tracking. This process is efficient as it simply requires the operator to move their head over the patient's torso. As the point cloud is captured, it is rendered in real-time through the headset for visual feedback. Overall, this process takes less than 10 seconds and conveniently fits into the workflow for human teleoperation.

To generate $N$, $\pmb{f}$, $W$, and $\pmb{\tau}$ for stiffness estimation, the US transducer pose and applied force and torque must be measured. We used a Philips Lumify US transducer (Philips, Amsterdam, Netherlands) instrumented with a 6-DOF force sensing shell \cite{black_force_2024} to measure the force and torque. To measure the pose, the NDI Polaris Spectra (Northern Digital Inc., Waterloo, Canada) was used to track an array of IR reflective spheres attached to the US transducer. Because the pose of the transducer must be measured relative to the patient model, additional ArUco markers were attached to the transducer to enable registration of the NDI Polaris Spectra to the Magic Leap 2. The fully instrumented transducer is shown in Figure~\ref{fig:probe}.

To evaluate the accuracy of the model, we conducted test scans on volunteer subjects ($n=4$) using the implementation described above. For each patient, three scanning trajectories were conducted: a vertical press, a sweep along the z-axis, and an oblique press. These trajectories were chosen based on observation of the motions sonographers conducted during abdominal aorta US exams. Each trajectory was performed twice for each patient so that one could be used to generate the force-informed model (we refer to this as the known trajectory), and the other to evaluate performance (we refer to this as the unknown trajectory). Data collection was performed at 20 Hz during these tests. Analysis of the data, including model generation and estimation of the rendered forces and torques, was completed offline in MATLAB. To quantify and compare accuracy, we calculated the mean absolute error (MAE), root mean squared error (RMSE) and root mean squared percent error (RMSPE) of the force and torque magnitudes, and the mean difference in force and torque vector angle (angle error).

\section{RESULTS}
\begin{figure*}[t]
    \centering
    \includegraphics[width=0.9\textwidth]{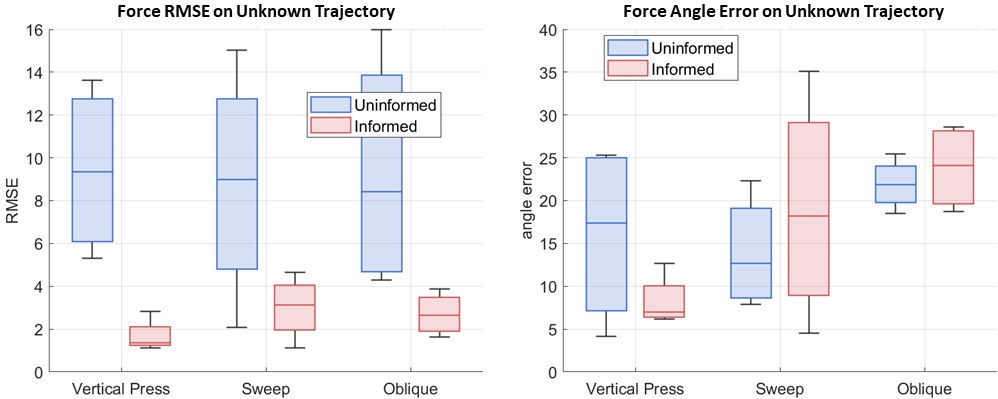}
    \caption{Box plots comparing distribution of force RMSE (N) and vector angle error ($^{\circ}$) between the force-informed model and the model without force information. The results presented in these plots are from the unknown trajectory.}
    \label{fig:forceBoxPlot}
\end{figure*}
\begin{figure*}[t]
    \centering
    \includegraphics[width=0.9\textwidth]{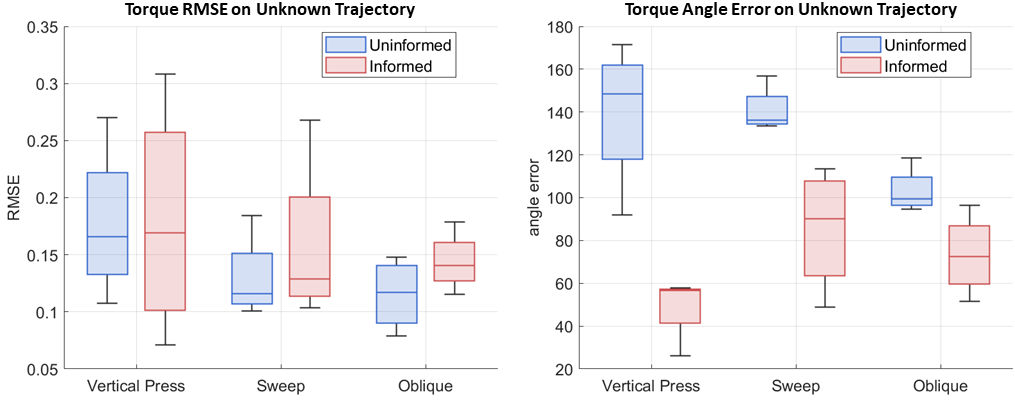}
    \caption{Box plots comparing distribution of torque RMSE (\SI{}{\newton\meter}) and vector angle error ($^{\circ}$) between the force-informed model and the model without force information. The results presented in these plots are from the unknown trajectory.}
    \label{fig:torqueBoxPlot}
\end{figure*}
\begin{figure*}[t]
    \centering
    \includegraphics[width=0.9\textwidth]{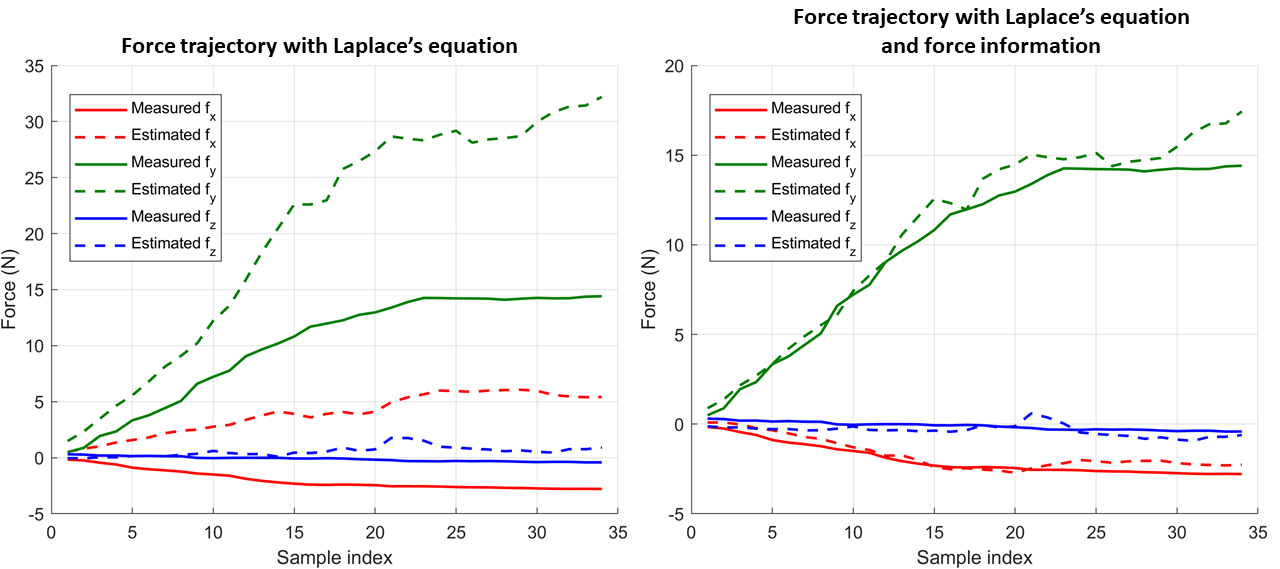}
    \caption{Force trajectory for sample 3 vertical press. There is a noticeable decrease in error from the trajectory without force information (left) to the trajectory with force information (right).}
    \label{fig:trajectory}
\end{figure*}
\begin{figure}[t]
    \centering
    \includegraphics[width=0.95\linewidth]{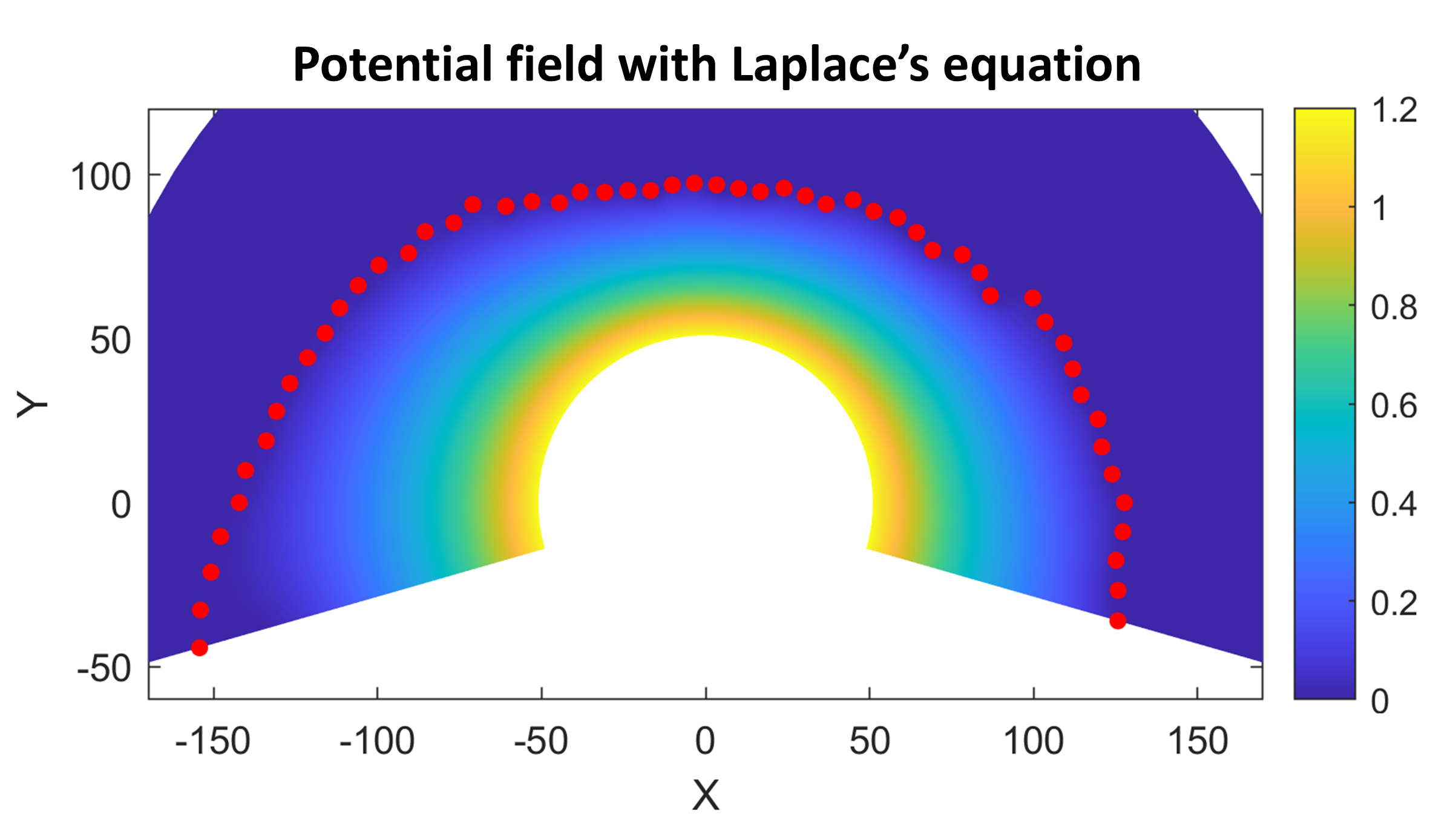} \\
    \includegraphics[width=0.95\linewidth]{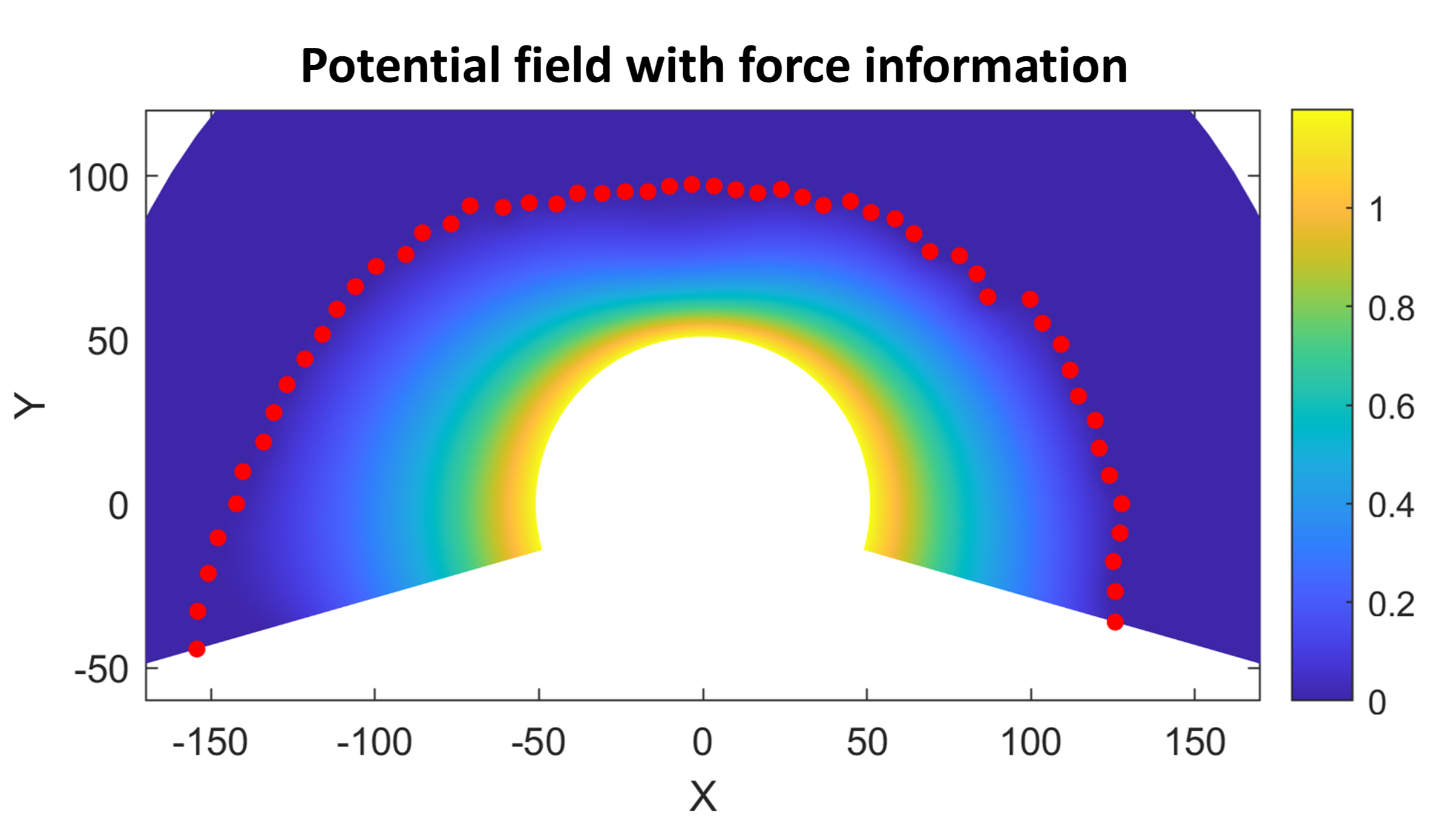} \\
    \includegraphics[width=0.95\linewidth]{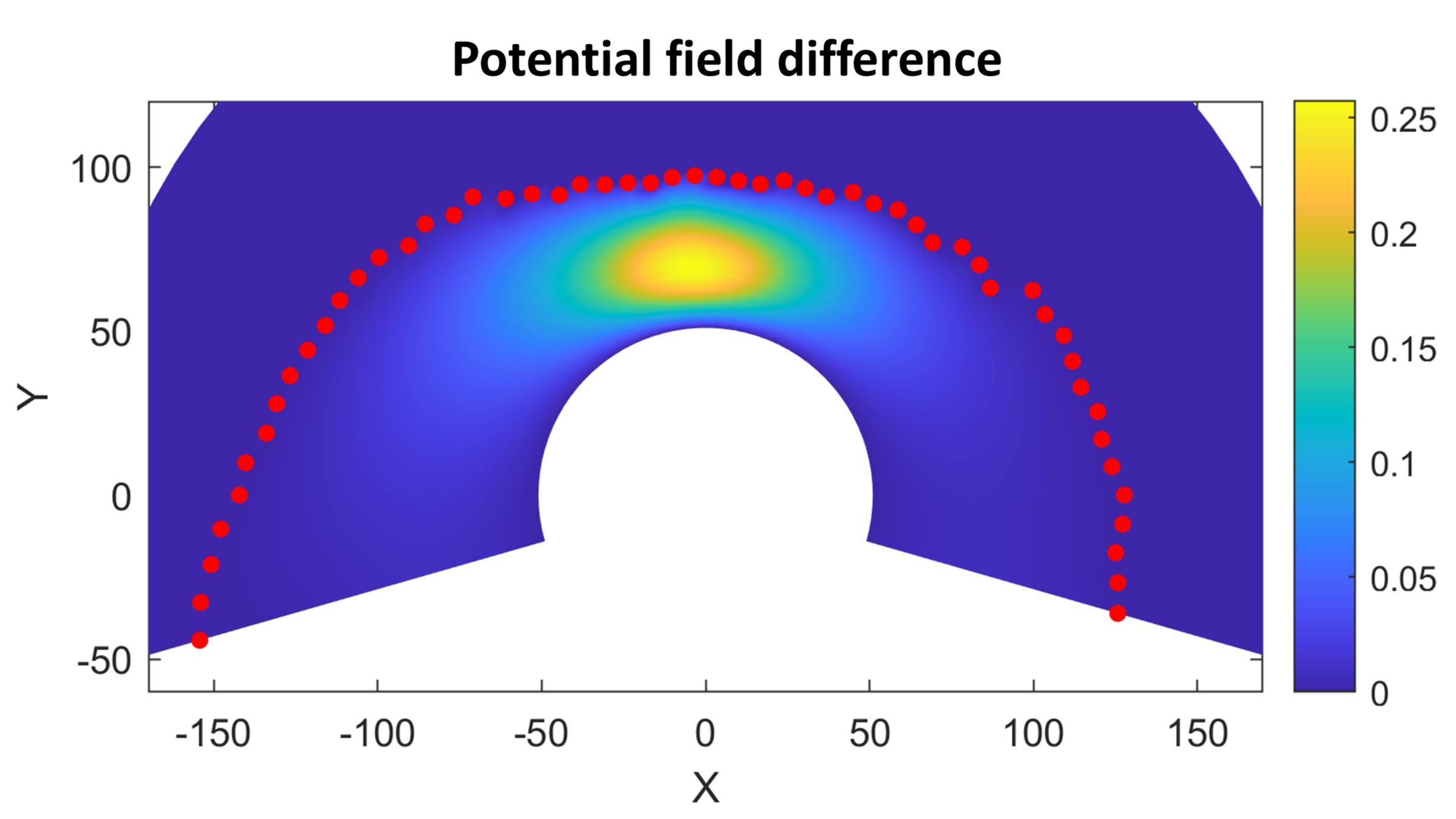}
    \caption{Heat map illustrating the potential field of a transverse slice of sample 3. Values outside the patient surface are set to 0 for clarity. Areas outside of the volume have no potential field values so are left white. The top figure shows the potential field without force information, the middle figure shows the force-informed potential field, and the bottom figure shows the difference between the two. Lighter regions indicate larger $p$ values and darker regions indicate smaller $p$ values. The surface point cloud is shown by the red dots.}
    \label{fig:potential}
\end{figure}

Table~\ref{tab:known_accuracy} presents the mean errors across all known trajectories and Table~\ref{tab:unknown_accuracy} presents the mean errors across all unknown trajectories. Additionally, the distributions of the force and torque RMSE and angle error for the unknown trajectories are illustrated in Figure~\ref{fig:forceBoxPlot} and Figure~\ref{fig:torqueBoxPlot}. On both the known and unknown trajectories, the force-informed model produced significantly smaller errors and standard deviations in force magnitudes. The force RMSE was reduced by a mean of $7.42\pm 4.47$~N on the known trajectories and $6.70\pm 4.49$~N on the unknown trajectories. This corresponds to a force RMSPE reduction of $65.5\pm 63.2\%$ on the known trajectories and $71.4\pm 78.6\%$ on the unknown trajectories. The force angle error was also reduced by a mean of $3.71\pm 10.5^\circ$ on the known trajectories and a mean of $0.24\pm 11.4^\circ$ on the unknown trajectories.

\begin{table}[h!]
    \centering
    \caption{Mean force and torque errors across all known trajectories, comparing the model without force information to the force-informed model.}
    \begin{tabular}{|c|c|c|}
        \hline
         & Laplace-only & Force-informed \\
        \hline 
        Force MAE (N) & $8.44\pm 4.19$ & $1.78\pm 0.82$ \\
        \hline 
        Force RMSE (N) & $9.57\pm 4.53$ & $2.15\pm 0.92$ \\
        \hline
        Force RMSPE (\%) & $94.4\pm 64.4$ & $29.0\pm 9.4$ \\
        \hline
        Force Angle Error ($^{\circ})$ & $17.1\pm 6.69$ & $13.4\pm 9.02$ \\
        \hline 
        Torque MAE (\SI{}{\newton\meter}) & $0.12\pm 0.05$ & $0.15\pm 0.06$ \\
        \hline 
        Torque RMSE (\SI{}{\newton\meter}) & $0.14\pm 0.05$ & $0.17\pm 0.07$ \\
        \hline
        Torque RMSPE (\%) & $87.7\pm 78.1$ & $95.9\pm 63.2$ \\
        \hline
        Torque Angle Error ($^{\circ}$) & $130\pm 29.9$ & $65.7\pm 30.9$ \\
        \hline
    \end{tabular}
    \label{tab:known_accuracy}
\end{table}

\begin{table}[h!]
    \centering
    \caption{Mean force and torque errors across all unknown trajectories, comparing the model without force information to the force-informed model.}
    \begin{tabular}{|c|c|c|}
        \hline
         & Laplace-only & Force-informed \\
        \hline 
        Force MAE (N) & $8.06\pm 4.24$ & $2.07\pm 1.04$ \\
        \hline 
        Force RMSE (N) & $9.15\pm 4.57$ & $2.46\pm 1.17$ \\
        \hline
        Force RMSPE (\%) & $103\pm 79.0$ & $31.9\pm 10.8$ \\
        \hline
        Force Angle Error ($^{\circ})$ & $17.3\pm 7.59$ & $17.0\pm 10.2$ \\
        \hline 
        Torque MAE (\SI{}{\newton\meter}) & $0.12\pm 0.05$ & $0.14\pm 0.07$ \\
        \hline 
        Torque RMSE (\SI{}{\newton\meter}) & $0.14\pm 0.05$ & $0.16\pm 0.07$ \\
        \hline
        Torque RMSPE (\%) & $105\pm 109$ & $112\pm 88.0$ \\
        \hline
        Torque Angle Error ($^{\circ}$) & $128\pm 26.8$ & $69.4\pm 25.1$ \\
        \hline
    \end{tabular}
    \label{tab:unknown_accuracy}
\end{table}

The force-informed model also produced significantly smaller torque angle errors across all trajectories. The torque angle error was reduced by a mean of $64.0\pm 42^\circ$ on the known trajectories and $58.5\pm 39.3^\circ$ on the unknown trajectories. However, the torque magnitude errors were mostly unchanged or slightly higher in the force-informed model.

Figure~\ref{fig:trajectory} shows an example trajectory comparison between the measured and rendered forces. It illustrates how the force-informed model results in a rendered force trajectory that is more aligned with the measured force trajectory. A single transverse slice of a potential field generated from sample~3 is visualized as a heat map in Figure~\ref{fig:potential}. A difference in the potential field can be seen around [0, 60] where forces were measured from the vertical press. Across the entire experiment, the mean time required to compute the force and torque for a single contact was $2.74$~ms for a point shell with $170$ points, which suggests force rendering can be done in real-time despite the computations being conducted offline in this experiment.

\section{DISCUSSION AND CONCLUSION}
We demonstrated our method of informing the Laplacian formulation with known forces and torques led to improved accuracy of the rendered forces on both the known and unknown trajectories. This is particularly evident in the significant reduction of force magnitude RMSE across all trajectories (Figure~\ref{fig:forceBoxPlot}). While the force angle error was clearly reduced in the force-informed model for the known trajectories, this was not the case for the unknown trajectories. This could be caused by the fact that the new trajectory visited some regions where forces were not measured in the known trajectory. If the points on the edge of the probe point shell were to fall into these regions, the force contributions from those points could shift the resulting force vector direction. Collecting more data that covers more of the patient's torso should improve the accuracy on new trajectories by minimizing the amount of unknown regions. In practice, we expect that as the sonographer scans the patient, position-force pairs of data are produced over the scanning region, thus identifying a full patient surface model with associated forces and torques.

The inclusion of measured torques led to significantly improved accuracy of the rendered torque direction on both the known and unknown trajectories. However, the torque magnitude accuracy in general did not show improvement. Part of this may be attributed to the lateral torques that were measured by the sensor during data collection whenever an axial force was applied to the US probe. Depending on how and where the US probe is held, it can be easy for the applied axial force to be slightly offset from the central axis which is aligned with the torque sensor coordinate origin. This would result in a large lateral torque measurement and can be difficult to capture in our model due to the minimal change in probe pose. Despite these results, the overall torques felt during scanning are relatively minimal compared to the forces, so the torque magnitude accuracy likely does not significantly affect the haptic experience. Regardless, further work will be necessary to investigate how the torque modelling can be improved. 

Another benefit of the force-informed model is that it becomes less sensitive to the selection of boundary conditions. Using sample 2 as an example, with Laplace's equation only, changing the interior boundary condition from $-0.1$ to $-1.2$ caused the force RMSE to change from $11.7$~N to $6.98$~N. However, with the force-informed model, the error changed by only $0.34$~N. Additionally, changing the boundary conditions did not show significant effect on the force angle error for either models.

While we have shown improvement in the force accuracy and the torque direction accuracy, further research is required to determine whether this would provide sonographers with a noticeable benefit during teleoperated US scans. The present work has already integrated the patient surface extraction with the human teleoperation system, including an efficient method of sending the structured patient point cloud. Future work will integrate the remaining components, including probe pose tracking and all of the model stiffness estimation, entirely on the human teleoperation system. This will allow teleoperated US scans to be performed through human teleoperation with patient modelling and stiffness estimation. A comparison can also be conducted with the proxy point cloud haptic rendering method described in \cite{ryden_proxy_2013}.

The methodology for force computation developed to present only includes quasi-static inter-penetration forces. In practice, we noticed there is some friction between the US transducer and the patient's skin that contributes to the overall error. Modelling and identification of such ``drag'' forces will be addressed in future work. The current model is also limited to stiffness estimation. Future work will investigate using the US transducer velocity to estimate and render damping associated with the internal potential field. To date, we have not carried out real-time implementations of the proposed stiffness estimation algorithm described in Section \ref{sec:estimation_algorithm}. We note that the force-informed implementation Laplacian formulation (\ref{eq:objective}) and its solution (\ref{eq:Qpd}-\ref{eq:d}) have straightforward recursive implementations. Each additional force/torque measurement has an effect on $Q$ which can be computed via rank-one updates through the Sherman-Morrison-Woodbury formula (Matrix Inversion Lemma).
\addtolength{\textheight}{-4cm}

The approach presented above for human teleoperation is identical to the approach that would be used for robotic teleoperation. A conventional force sensor or the force sensor used therein would enable force and torque measurement during scanning, and the trajectories would be available from the robot forward kinematics. Instead of the TOF depth camera of a mixed reality headset like the Magic Leap 2, there are many other depth cameras with suitable range, such as the Intel RealSense (Intel Corporation, California, USA) with objective choice to match the average distance to the patient.

\bibliography{IEEEabrv,IEEEexample}





\end{document}